\documentclass{article}
\usepackage{microtype}
\usepackage{graphicx}
\usepackage{subcaption}
\usepackage{booktabs} 
\usepackage{enumitem}
\usepackage{multirow}
\usepackage{arydshln}
\usepackage{hyperref}
\newcommand{\benchmark}{{MIMIC-ED-Assist}}
\newcommand{\method}{{ED-Copilot}}

\usepackage[accepted]{icml2024}

\usepackage{amsmath}
\usepackage{amssymb}
\usepackage{mathtools}
\usepackage{amsthm}

\usepackage[capitalize,noabbrev]{cleveref}

\theoremstyle{plain}

\theoremstyle{definition}

\theoremstyle{remark}

\usepackage[textsize=tiny]{todonotes}

\definecolor{midnightgreen}{rgb}{0.0, 0.29, 0.33}
\definecolor{brickred}{rgb}{0.8, 0.25, 0.33}
\definecolor{desert}{rgb}{0.76, 0.6, 0.42}

\icmltitlerunning{ED-Copilot: Reduce Emergency Department Wait Time with Language Model Diagnostic Assistance}

\begin{document}

\twocolumn[

\icmltitle{ED-Copilot: Reduce Emergency Department Wait Time \\
with Language Model Diagnostic Assistance}

\icmlsetsymbol{equal}{*}

\begin{icmlauthorlist}
\icmlauthor{Liwen Sun}{CMU}
\icmlauthor{Abhineet Agarwal}{BerkeleyStats}
\icmlauthor{Aaron Kornblith}{UCSF}
\icmlauthor{Bin Yu}{BerkeleyStats,BerkeleyEECS}
\icmlauthor{Chenyan Xiong}{CMU}
\end{icmlauthorlist}


\icmlaffiliation{CMU}{Language Technologies Institute, Carnegie Mellon University} 
\icmlaffiliation{BerkeleyStats}{Department of Statistics, University of California, Berkeley}
\icmlaffiliation{BerkeleyEECS}{Department of EECS, University of California, Berkeley}
\icmlaffiliation{UCSF}{Department of Emergency Medicine \& Pediatrics, University of California, San Francisco}

\icmlcorrespondingauthor{Chenyan Xiong}{cx@cs.cmu.edu}

\icmlkeywords{AI for Health Care; Language Models for Emergency Department}
\vskip 0.3in
]

\printAffiliationsAndNotice{}

\begin{abstract}
In the emergency department (ED), patients undergo triage and multiple laboratory tests before diagnosis. 
This time-consuming process causes ED crowding which impacts patient mortality, medical errors, staff burnout, etc.
This work proposes (time) \textit{cost-effective diagnostic assistance} that leverages artificial intelligence systems to help ED clinicians make efficient and accurate diagnoses. 
In collaboration with ED clinicians, we use public patient data to curate \benchmark, a benchmark for AI systems to suggest laboratory tests that minimize wait time while accurately predicting critical outcomes such as death. 
With \benchmark, we develop \method\ which sequentially suggests patient-specific laboratory tests and makes diagnostic predictions.
\method\ employs a pre-trained bio-medical language model to encode patient information and uses reinforcement learning to minimize ED wait time and maximize prediction accuracy. 
On \benchmark, \method\ improves prediction accuracy over baselines while halving average wait time from four hours to two hours. 
\method\ can also effectively personalize treatment recommendations based on patient severity, further highlighting its potential as a diagnostic assistant. 
Since \benchmark\ is a retrospective benchmark, \method\ is restricted to recommend only observed tests. 
We show \method\ achieves competitive performance without this restriction as the maximum allowed time increases. 
Our code is available at \url{https://github.com/cxcscmu/ED-Copilot}.

\end{abstract}

%



\section{Introduction}

Emergency Department (ED) crowding represents a critical challenge in healthcare, significantly impacting morbidity, mortality, medical error, staff burnout, and incurring excessive costs \cite{Sartini2022}. 
Despite the documented effects of ED crowding, this issue remains inadequately addressed in healthcare systems. 
An efficient and effective ED is vital for providing timely care to severely ill or injured patients \cite{Savioli2022EDOvercrowding}.

One key area to address ED crowding, as identified by the American College of Emergency Physicians, is to enhance \textit{throughput}---the efficacy and efficiency of care delivery in the ED~\citep{Jarvis2016EDPatientFlow, DEANDA2018582}.
A crucial factor affecting throughput is the laboratory testing process, where patients often face lengthy waits for tests to be ordered and completed, delaying diagnoses and treatment decisions ~\citep{li2015effect}. 
Studies also show that 40 to 60\% of ED laboratory tests are unnecessary~\cite{miyakis2006factors}, further exacerbating wait times.

This paper proposes an artificial intelligence ``Co-Pilot'' system intended to offer (time) cost-effective diagnostic assistance in the ED. 
This system should aid diagnosis and minimize ED length of stay (LOS), i.e., wait times, by suggesting laboratory tests after patient triage. 
Further, it should help with resource management and planning by identifying severely ill patients who require rapid intervention.
That is, by selecting informative tests, the system streamlines the diagnostic process, reducing LOS while improving outcomes, particularly for high-risk patients. 


To support the machine learning (ML) community in developing a time-cost-effective diagnostic assistant, we collaborate with ED clinicians to curate a benchmark, called \benchmark, that is derived from MIMIC-IV \citep{johnson2023mimic} and related datasets \cite{xie2022benchmarking}. 
\benchmark\ is designed to test the ability of AI systems to provide both accurate and time-cost saving laboratory recommendations.  
Our benchmark consists of two prediction targets identified by our clinical collaborators to reflect patient risk: critical outcomes which includes patient death and ICU transfer~\citep{LEVIN2018565}, and lengthened ED stay, defined as ED LOS exceeding $24$ hours. 
Accurately identifying patients at high risks of these outcomes reduces time-cost by allowing clinicians to perform timely interventions and efficiently allocate resources. 
\benchmark\ mirrors real-world ED practices by grouping individual laboratory tests into commonly performed groups, e.g., complete blood count (CBC). 
MIMIC-ED-Assist then tests AI systems on their ability to recommend the most informative groups to make accurate diagnostic suggestions while minimizing the total time required to perform these tests, thereby reducing LOS.

With \benchmark, we propose \method\ which suggests a series of laboratory groups to flag patients at high risks on our prediction targets while minimizing total time-cost. 
\method\ first linearizes (i.e., converts to text) patient information, including demographic, triage, and laboratory test results into a text sequence. 
It then fine-tunes a bio-medical pre-trained language model BioGPT~\cite{luo2022biogpt} to suggest future groups and predict our two defined targets.
Next, we use a reinforcement learning (RL) framework \citep{yu2023deep} to teach BioGPT to dynamically recommend the subsequent, most informative laboratory group based on prior laboratory and triage information.
Unlike baselines, \method~serves as a personalized diagnostic assistant since it uses past patient information to  recommend future medically relevant laboratory groups. 

Experiments on \benchmark\ show that \method\ outperforms state-of-the-art tree models while halving time-costs of laboratory testing from four hours to two hours. 
Reducing the number of laboratory tests also has the benefit of reducing financial cost. 
We also perform ablation studies to confirm the benefits of our feature linearization technique and the bio-medical pre-trained language model backbone.
Our ablation studies also investigate the effect of size of the language model backbone on prediction accuracy which indicates larger models can lead to further gain in performance. 

Our analyses also confirm the benefit of \method's personalized modeling approach. 
We show \method\ can adapt its recommendations based on patient severity, thereby providing more accurate diagnostic suggestions for severely ill patients as compared to non-personalized baselines. 
Further, \method\ achieves consistent performance across various subgroups such as age and sex. 
Lastly, since \benchmark\ is a retrospective offline benchmark, we restrict \method\ to only select laboratory tests a patient actually receives. 
We perform simulations without this restriction to approximate online performance, and show \method\ is still able to make medically appropriate recommendations. 

The rest of this paper is organized as follows. In \cref{sec:related_work}, we review related work. 
We discuss \benchmark\ and \method\ in Sections \ref{sec:benchmark_descripton} and \ref{sec:ed_copilot} respectively. 
Sections \ref{sec:experimental_set_up} and \ref{sec:results} discuss our experimental set-up and results.

\section{Related Work}
\label{sec:related_work}

\textbf{Healthcare Benchmarks.}  Researchers have spent considerable effort in converting raw electronic health records (EHRs) into large-scale open-source datasets to ensure easy access to high-quality medical data. A notable example is the Medical Information Mart for Intensive Care (MIMIC) database \cite{johnson2023mimic} which provides patient information such as measurements, laboratory orders, and treatments, ranging from the ED to inpatient care, including the intensive care unit (ICU). MIMIC has led to the development of a range of related prediction benchmarks and models \cite{PURUSHOTHAM2018112,Harutyunyan_2019,Wang_2020} focused on the ICU. \citet{xie2022benchmarking} took a step towards filling this gap by using the MIMIC-IV-ED \cite{Johnson2023MIMICIVED} database to build a ED-focused benchmark. Their dataset includes ED triage information, and various clinical outcomes such as hospitalization that interest clinicians, and impact ED LOS.

\textbf{AI Models for Healthcare.} There has been significant effort to apply ML to accurately predict clinical outcomes.  
Traditional methods (e.g., random forests, gradient boosting, and their variants \cite{breiman2001random,chen2016xgboost,agarwal22bHS,agarwal2023mdi}), along with deep learning (DL) have been used to predict pneumonia \cite{kang2020predicting}, and septic shock in the ICU \cite{wardi2021predicting}.
Other works use interpretable models to provide diagnostic assistance in the ED such as identifying traumatic brain injury~\cite{kornblith2022predictability,tan2022fast}.
Another closely related line of work is AI for cost-effective medicine.
For instance, \citet{bejnordi2017diagnostic} showed DL led to faster analysis of pathology laboratory results; \citet{komorowski2018artificial} proposed an ``AI clinician'' to learn optimal dosing strategies for treating sepsis; \citet{yu2023deep} focused on minimizing financial costs associated with laboratory testing while maximizing prediction accuracy.
Specifically, \citet{yu2023deep} used RL to sequentially select laboratory groups based on a patient's observed test results to optimize this (financial) cost-accuracy trade-off. 
They validated the accuracy and cost-effectiveness of their approach on multiple clinical tasks such as diagnosing kidney injury. 
Researchers have also begun to explore the use of large language models (LLMs) in medical applications. 
LLMs have been used to extract clinical concepts \cite{yang2021clinical, luo2022biogpt,Yang2022LargeLanguageModelEHR}, and facilitate medical question answering \cite{singhal2023expertlevel,yagnik2024medlm}.
%


\section{\benchmark~Benchmark}
\label{sec:benchmark_descripton}
In this section, we discuss the curation of \benchmark\ in collaboration with ED clinicians.

\textbf{Task Description.} We consider the following two tasks relevant to reducing ED LOS. 1) Flagging patients at high risks of critical outcome (i.e., death and ICU transfer), and ED LOS exceeding $24$ hours. 
2) Providing time-cost-effective diagnostic assistance by recommending the next most medically informative laboratory group while simultaneously minimizing the time-cost of these groups. 

\textbf{Data Pre-processing.}  Since laboratory results are only available for admitted patients, we filter out non-hospitalized ED patients from MIMIC-IV-ED.
We only focus on adults and remove patients younger than $18$ years old. 
We also exclude patients that miss triage information.
This step is necessary since clinicians order laboratory groups depending on triage information.
To simplify the task, we remove patients who receive the same test multiple times, approximately $1.5\%$ of all patients.

\textbf{Clinically Relevant Outcomes.} In collaboration with ED clinicians, we chose the two following prediction targets.
(1) Critical outcome, which refers to death during hospitalization or transfer to an ICU within 12 hours.
Identifying patients at high risks of critical outcome allows clinicians to prioritize treatment and resources for them. 
(2) Lengthened ED stay, indicating if ED LOS exceeding $24$ hours.
Lengthened ED stay is typically correlated with the complexity of a patient's case. 
Flagging patients at high risks of lengthened ED stay can enable timely intervention, and reduce ED LOS. 
The proportion of patients with these outcomes is described in Table \ref{tab:dataset_summary_statistics}.
While these two outcomes are correlated, they also cover different aspects of patient care.
For example, patients at high risks of critical outcomes often should be hospitalized quickly, while patients with complications do not necessarily have severe cases.
As such, healthcare providers often require different diagnostics and resource managements for these two tasks.

\textbf{Triage Feature Selection.} Triage features are measurements that are available for every patient before laboratory tests are ordered. 
We select a number of triage features in collaboration with ED clinicians.
Specifically, we chose $9$ triage variables available at the beginning of patient encounters, which include patient demographics, medical history, vital signs, and chief complaints (i.e., natural language description of symptoms). 

\textbf{Laboratory Test Selection.} We only include laboratory tests performed in the ED. 
For simplicity, we exclude tests received by less than $5\%$ of patients and leave examination of rare tests to future research.
This process results in a total of $68$ available laboratory tests.
While there are $68$ tests, ED clinicians rarely order individual tests for a patient. 
Typically, they order groups of tests (e.g., complete blood count) based on a patient's signs, symptoms, and risk factors.
To reflect this clinical practice, our clinical collaborators categorized these 68 tests into 12 distinct groups. 
See \cref{sec:triage_lab_features} for all $68$ tests, and their assigned groupings.
On average each patient receives $4.7$ laboratory groups.
Consequently, \benchmark\ contains numerous missing values, each representing a laboratory test not administered to a specific patient.

\textbf{ED LOS.} Our benchmark records when each group is ordered, and assigns its average completion time as its 'time-cost'. 
ED LOS depends on the time-costs of the administered groups and can be modeled in different ways. 
For example, sequential tests result in ED LOS being the sum of time-costs, whereas parallel tests imply ED LOS is equal to the group with the largest time-cost. 
\benchmark\ does not specify how to approximate ED LOS from these time-costs, and instead provides this flexibility to researchers and practitioners. 

\begin{table}[t]
 \caption{
 Statistics of \benchmark. It includes information from patient triage and laboratory tests. Laboratory tests are grouped into 12 groups by ED clinicians based on how they are commonly ordered. An ED visit has a critical outcome if the patient is transferred to ICU or there is an inpatient mortality. ED stay is lengthened if the length of stay (LOS) exceeds 24 hours. The positive rate is shown in parentheses.
 }  
   \vspace{2mm}
    \label{tab:dataset_summary_statistics}
    \centering
    \small
    \begin{tabular}{l r} 
    \toprule
    \textbf{Variable/Label} & \textbf{Count} \\
    \midrule
        $\#$ of ED visits & 32356\\
        $\#$ of patients & 25714\\
        $\#$ of triage variables &9 \\
        $\#$ of laboratory variables & 67\\
        $\#$ of laboratory groups & 12\\
       Avg. $\#$ of laboratory groups per patient &4.7 \\
    \midrule
        $\#$ of Inpatient mortality& 467 (1.44\%)  \\
        $\#$ of ICU transfer in 12h & 2894 (8.94\%) \\
        $\#$ of Critical outcome &3129 (9.67\%) \\
    \midrule
        $\#$ of ED LOS $>$ 24h & 2232 (6.90\%)\\
    \bottomrule
    \end{tabular}
   \end{table}


\textbf{Data Availability.} Our pipeline to create \benchmark\ from the MIMIC-IV dataset can be found at \url{https://github.com/cxcscmu/ED-Copilot}. After completing a training course and signing a data use agreement regarding patient information privacy, individuals will gain access to MIMIC-IV and can utilize our pipeline to create \benchmark.


\textbf{Limitations.}  Since \benchmark\ is derived from public patient data, it suffers from some downsides due to a lack of data availability.  
(1) \benchmark\ is derived from MIMIC-IV which only has laboratory results for hospitalized ED patients.
Thus, our dataset is not reflective of the entire population that visits the ED, but instead is biased towards those with more severe issues (hospitalized).
(2) As an offline benchmark derived from past patient records, all data is retrospective. 
As a result when developing diagnostic assistant, one can only use laboratory tests patients actually received, otherwise no testing results are available. 
This leads to common challenges of offline benchmarks and may result in models learning sub-optimal recommendations, and unobserved confounding. 
We note that measuring `online' performance of \method\ and AI systems in general will require clinical trials which is out of the scope of this paper. 
(3) MIMIC-IV is collected at the Beth Israel Deaconess Medical Center, and may not reflect the distribution of other healthcare systems. 
Caution is needed when generalizing models and insights derived from \benchmark\ to new healthcare systems.

%

%
%

\section{ED-Copilot for Diagnostic 
Assistance}
\label{sec:ed_copilot}

\begin{figure}[t]
\centering
	\includegraphics[width =\textwidth/2]{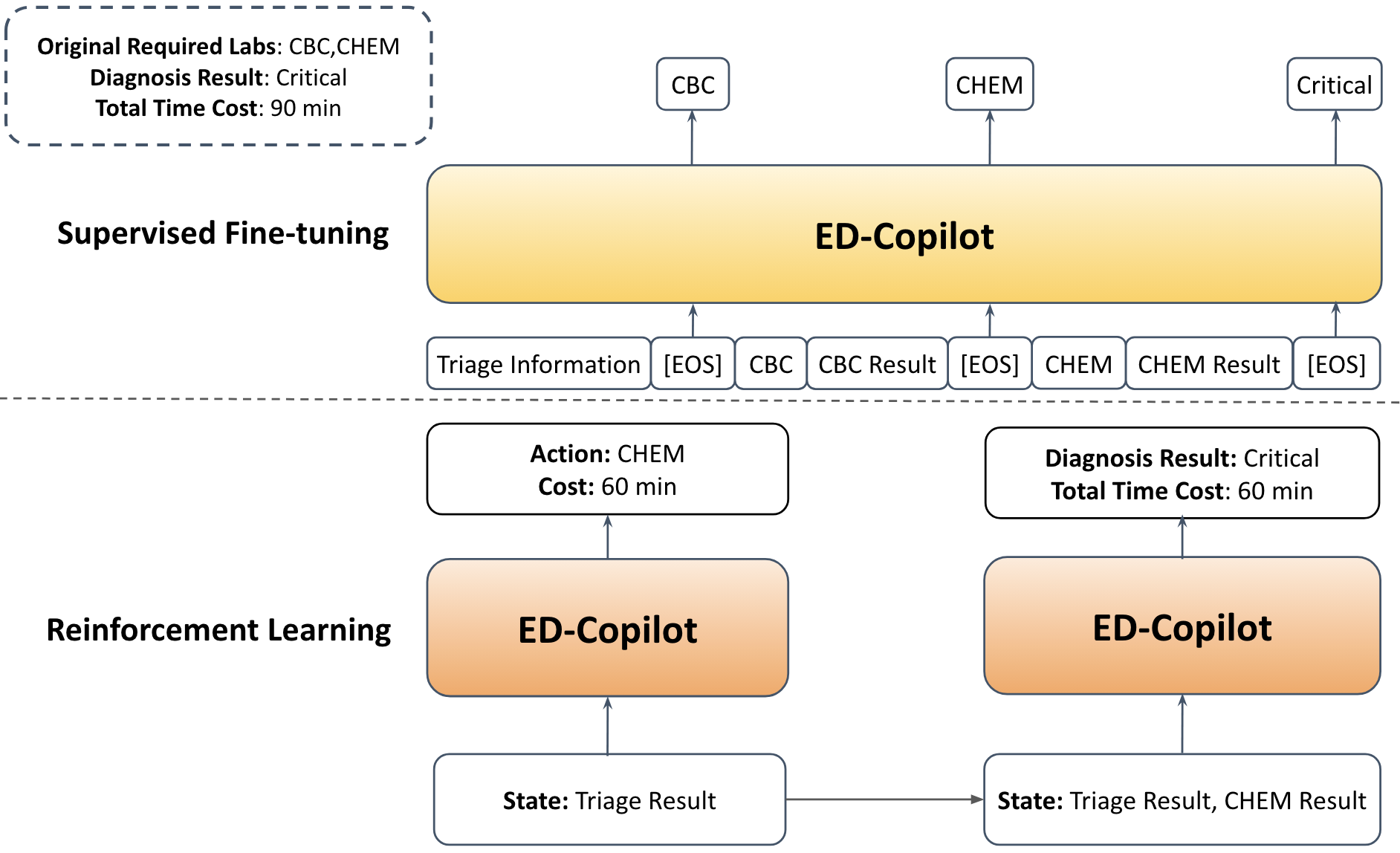}
	\caption{Overview of \method\ procedure on one ED visit.}
	\label{fig:overview}
\end{figure}

We use \benchmark\ to develop \method, which offers cost-effective diagnostic assistance by minimizing the number of laboratory groups required to identify high-risk patients. 
This section is organized as follows. 
First, we provide a high-level overview of \method, see Figure \ref{fig:overview} for a visualization. 
Then, we detail its training process which consists of two stages: supervised fine-tuning to adapt the pre-trained language model (PLM) (e.g., BioGPT) to our prediction task, followed by RL to select laboratory groups that reduce time-costs. 
Finally, we discuss how \method\ conducts inference. 

\subsection{Problem Formulation}
\label{subsec:ed_copilot_architecture}

Let a patient has triage features denoted by $x_0$ and $n$ laboratory groups  $ [x_i]_{i=1}^n$with their associated results $ [r_i]_{i=0}^n$ in order $[x_0,r_0,x_1, r_1,x_2,r_2,....,x_n,r_n]$. 
Additionally, let $y$ denotes our prediction targets defined in \cref{sec:benchmark_descripton}.
ED-Copilot first linearizes this patient's triage and laboratory results into a text sequence, which is then inputted into the PLM backbone to suggest the next group of tests or predict the outcome. 
We describe these steps in detail as follows. 

\textbf{Linearization.} As laboratory results and triage information are stored in a tabular format, we first convert this information to text for a PLM to use. 
Specifically, we linearize laboratory group results via the following text template \citep{hegselmann2023tabllm}  $r = \textit{test name} :  \textit{test value} |  \textit{test name} :  \textit{test value} ...$
\textit{Test name/value} refers to the name and recorded result of the ordered laboratory test respectively. 
Each \textit{test name/value} pair is separated by a pipe ($|$).

\textbf{Language Model Backbone.}  We apply PLM $G_{\theta}$ to the text sequence constructed above to obtain hidden representations $\textit{\textbf{H}}$ for each [\texttt{EOS}] token. 
We then use a MLP $p_{\phi}(x_{i}|\textbf{\textit{h}}_{< i})$ to predict the next laboratory group using hidden representations $\textbf{\textit{h}}_{< i}$ for tokens \texttt{[EOS]}$_{<i}$. 
Lastly, we predict outcomes $y$ using $ p_{\psi}(y| \textit{\textbf{h}}_{\leq n} )$ where $\textit{\textbf{h}}_n$ is the hidden state for token \texttt{[EOS]}$_{n}$.
This process is reflected below. 
\begin{align}
    [ x_0, r_0,\texttt{[EOS]}_0,&...,    x_n,r_n,\texttt{[EOS]}_n,y ] 
    \xrightarrow{G_{\theta}} \textit{\textbf{H}}, \\
     p_{\phi} (x_i | \textit{\textbf{h}}_{<i} ) &= \text{MLP}_{\phi}(\textit{\textbf{h}}_{i-1} ), \\
    p_{\psi}(y| \textit{\textbf{h}}_{\leq n} ) &= \text{MLP}_{\psi} (\textit{\textbf{h}}_n ).
\end{align}
Note that we only predict laboratory groups, not their associated laboratory tests' results, which should be determined by conducting the actual laboratory tests. Next, we describe our training procedure. 

\subsection{Supervised Fine-tuning}
\label{subsec:fine_tuning}
In this section, we perform supervised fine-tuning of our PLM to suggest the next laboratory group and predict outcomes.
To predict the next laboratory group, we use a standard auto-regressive loss function,
\begin{align}
    \mathcal{L}_{\text{lab}} & = -\frac{1}{n}\sum_{i=1}^{n} \log p_{\phi} (x_i | \textit{\textbf{h}}_{<i}).    
\end{align}
For outcome prediction, we use the following loss,
\begin{align}
    \mathcal{L}_{y}  & = -\log p_{\psi} (y| \textit{\textbf{h}}_{\leq n} ). 
\end{align}
To fine-tune, we minimize the loss, 
\begin{align}
\label{eq:loss_supervised}
     \theta^*,\psi^*,\phi^* = \min_{\theta,\psi,\phi}  ( \mathcal{L}_{\text{lab}} +  \mathcal{L}_y ),
\end{align}
where $\theta,\psi,\phi$ are parameters in the PLM $G_{\theta}$ and two MLPs $p_{\psi},p_{\phi}$. 
ED clinicians can use the fine-tuned PLM to suggest a sequence of laboratory groups and predict outcomes. 
%
%
\subsection{Reinforcement Learning}
\label{subsec:rl}
Choosing informative laboratory groups is a key factor in reducing laboratory testing, and thereby reducing ED LOS. 
In this section, we employ RL to introduce the notion of time-cost effectiveness to the fine-tuned PLM to select laboratory groups that maximize predictive accuracy while minimizing time-cost.

\textbf{Markov Decision Process.} 
The RL process can be viewed as a Markov Decision Process (MDP),  represented by $\{\mathcal{S},\mathcal{A},P,R\}$. 
Let $\mathcal{S}$ denote the state space, and state $s \in \mathcal{S}$ corresponds to a patient's observed triage information, and laboratory group results. 
Specifically, for a patient with observed information $[x_0,r_0,x_1, r_1,x_2,r_2,....,x_n,r_n]$, let $s_{\leq i} = [x_0,r_0,x_1, r_1,x_2,r_2,....,x_i,r_i]$ for $0\leq i \leq n$.

\textbf{Action Space.} Denote the action space as $\mathcal{A} = \{x_1,x_2,....,x_K\} \cup \{y^+, y^-\}$. Action $a \in \{x_1,x_2,....,x_K\}$ corresponding to ordering a group of laboratory tests with associated time-cost $c(a)$. Action $a \in \{y^+, y^-\}$ refers to predicting an outcome and terminating the MDP.

\textbf{Policy.} The policy $\pi_{\eta}: \mathcal{S} \rightarrow \mathcal{A}$ maps from states to actions, and is parameterized by $\eta$.
Specifically, $\pi_{\eta}(a|\textbf{\textit{h}}_{< i})$ outputs  probability of an action $a$ using the hidden representation $\textbf{\textit{h}}_{< i}$  corresponding to \texttt{[EOS]} token for state $s_i$.

\textbf{RL Training.} We train the policy $\pi_{\eta}$ to follow two objectives: maximize F1-score and minimize time-cost. 
We measure the time-cost of policy $\pi_{\eta}$, as follows: 
\begin{align}
\label{eq:policy_cost_def}
\text{Cost}(\pi_{\eta}) =  \mathbb{E}_{\pi_{\eta}}[\sum_{t \geq 0}\sum_{j \in  [K]}c(j)\cdot \mathbf{1}\{a_t = x_j\}],
\end{align}
where $\mathbb{E}_{\pi_{\eta}}$ is the expectation under policy $\pi_{\eta}$.
Using \eqref{eq:policy_cost_def}, let $\pi^{*}_{\eta}(\alpha,\beta)$ represent the policy that maximizes F1-score while minimizing time-cost for hyper-parameters $\alpha,\beta$ to be defined. 
Then, $\pi_{\alpha,\beta}^{*}$ is equivalent to the solution of the following program:
\begin{equation}
\label{eq:optimal_policy}
\begin{split}
     \pi^{*}_{\eta}(\alpha,\beta)= \text{argmax}_{\pi_\eta}\{&\text{TN}(\pi_\eta) +\alpha \text{TP}(\pi_\eta)\\
     &+ \beta\text{Cost}(\pi_\eta)  \}.
\end{split}
\end{equation}
TN$(\pi_{\eta})$ and TP$(\pi_{\eta})$ are the numbers of true negatives and true positives under policy $\pi_{\eta}$. Hyper-parameters $\alpha,\beta$ control the trade-off between F1-score and time-cost. 
We train the MLP via proximal policy optimization \cite{schulman2017proximal}.

\textbf{Measuring Time-cost.} We measure time-cost via the total time taken to run all laboratory groups.
In the ED, laboratory groups are often ordered both in parallel, and sequentially since the decision on new tests to order depends on previous tests. 
Additionally, the number of tests an ED clinician orders depends on other factors outside of the patient's health record, such as insurance policy. 
As the first step towards reducing ED LOS with AI, we use the sum of laboratory group time-costs as the total cost, which serves as an approximation to the effect of testing on ED LOS.   
Better ways to model ED LOS are future research for both the AI and healthcare communities.

\begin{table*}[h]
\caption{Overall performance of ED-Copilot and baselines in predicting \textit{Critical Outcome} (transfer to ICU or mortality) and \textit{Lengthened ED Stay} (exceeding 24 hours in the ED), as well as the \textit{Average Time-cost} (in minutes) to perform laboratory tests. Sensitivity and specificity are true positive and true negative rates. We report results averaged over three random seeds alongside standard deviations. }
\label{major}
\centering
\resizebox{\textwidth}{!}{
    \begin{tabular}{lccccrccccr}
    \toprule
    & \multicolumn{5}{c}{\textbf{Critical Outcome}} & \multicolumn{5}{c}{\textbf{Lengthened ED Stay}} \\
    \cmidrule(lr){2-6} \cmidrule(lr){7-11}
    \textbf{Model} & F1 & AUC & Sensitivity & Specificity & Avg. Time-cost   & F1 & AUC & Sensitivity & Specificity & Avg. Time-cost \\
    \midrule
Random Forest          & 0.377 (0.015)  & 0.807 (0.011) & 0.754 (0.012)   & 0.748 (0.005)   & 265 Min           & 0.206 (0.014)  & 0.698 (0.011) & 0.693 (0.016)   & 0.616 (0.024)   & 265 Min          \\ 
XGBoost                & 0.379 (0.019)  & 0.807 (0.009) & 0.731 (0.017)   & 0.744 (0.006)   & 265 Min        & 0.212 (0.010)  & 0.679 (0.007) & 0.619 (0.020)   & 0.661 (0.020)   & 265   Min        \\ 
LightGBM               & 0.394 (0.016)  & 0.813 (0.008) & 0.725 (0.012)   & 0.769 (0.004)   & 265 Min         & 0.217 (0.015)  & 0.705 (0.011) & 0.706 (0.017)   & 0.605 (0.014)   & 265  Min         \\ 
3-layer DNN            & 0.339 (0.032)  & 0.743 (0.021) & 0.676 (0.024)   & 0.683 (0.011)   & 265 Min         & 0.194 (0.031)  & 0.637 (0.013) & 0.649 (0.015)   & 0.593 (0.014)   & 265  Min         \\ 
\midrule
SM-DDPO                & 0.353 (0.031)  & 0.780 (0.020) & 0.685 (0.023)   & 0.763 (0.022)   & 182 (32) Min      & 0.183 (0.028)  & 0.619 (0.012) & 0.472 (0.012)   & 0.739 (0.011)   & 177 (60) Min     \\ 
ED-Copilot             & \textbf{0.413 (0.028)}  & \textbf{0.820 (0.021)} & \textbf{0.750 (0.018)}   & \textbf{0.779 (0.011)}   & \textbf{125 (21)} Min     & \textbf{0.232 (0.023)}  &\textbf{ 0.707 (0.015)} & \textbf{0.725 (0.018)}   & \textbf{0.606 (0.015)}   & \textbf{154 (33)} Min      \\ 
    \bottomrule
    \end{tabular}
}

\end{table*}

\subsection{Inference}
\label{subsec:infer}
During inference, \method~assists clinicians to optimize their workflow by suggesting the next most informative laboratory group, and also by flagging patients at high risks of critical outcome and lengthened ED stay.
Specifically, given triage information and previous test results, \method~recommends additional tests. 
The results of these tests are fed back to \method~to suggest additional laboratory groups or to flag one of two possible outcomes. 
Further, \method~can be used in more others ways depending on clinical needs. 
For example, multiple tests can be ordered using multiple suggestions from \method{}.

\begin{figure*}[h]
    \centering
    \begin{subfigure}[t]{0.25\textwidth}
        \centering
        \includegraphics[width=\textwidth]{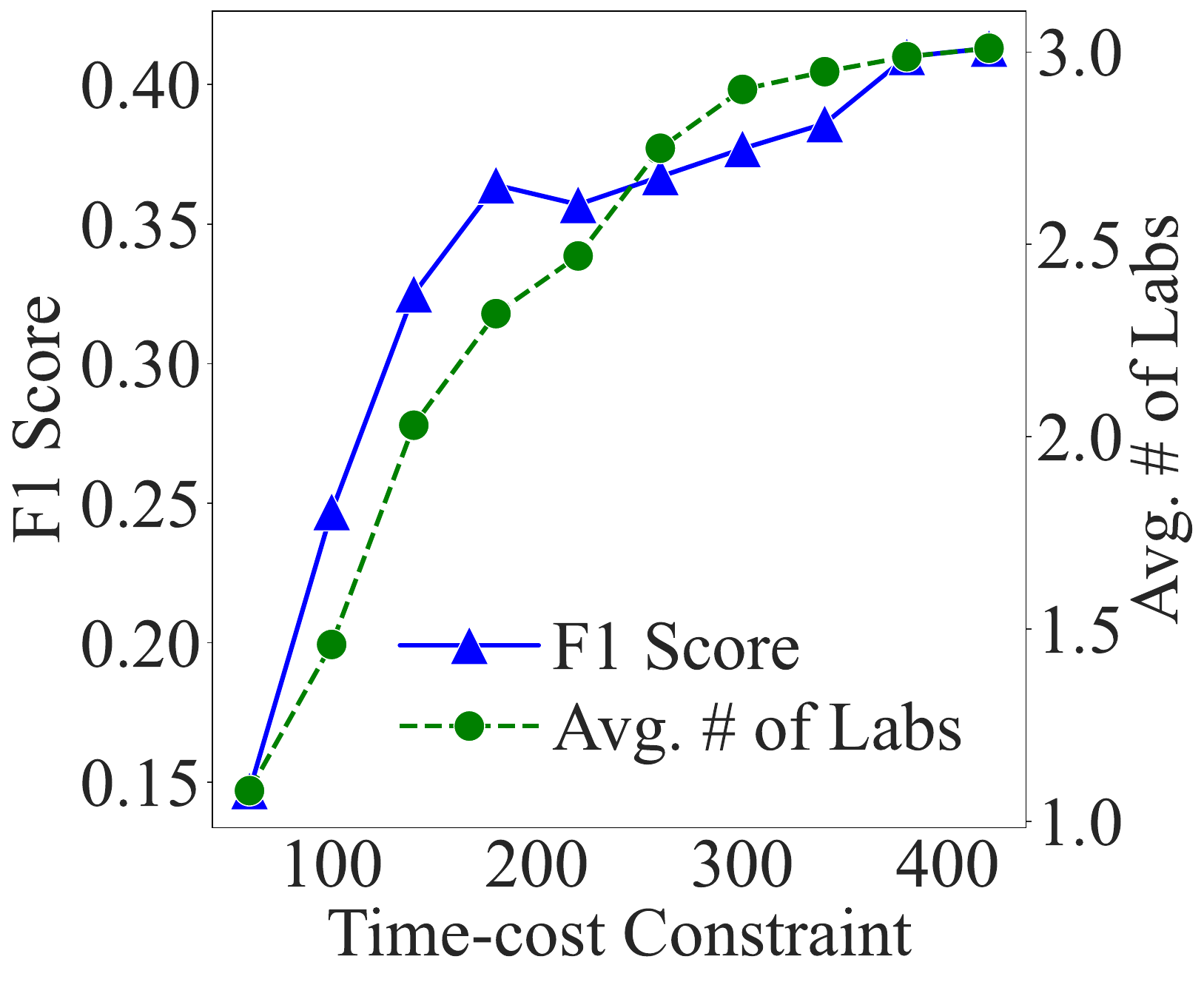}
        \caption{Critical Outcome F1}
        \label{fig:curve_critical_f1}
    \end{subfigure}%
    \begin{subfigure}[t]{0.25\textwidth}
        \centering
        \includegraphics[width=\textwidth]{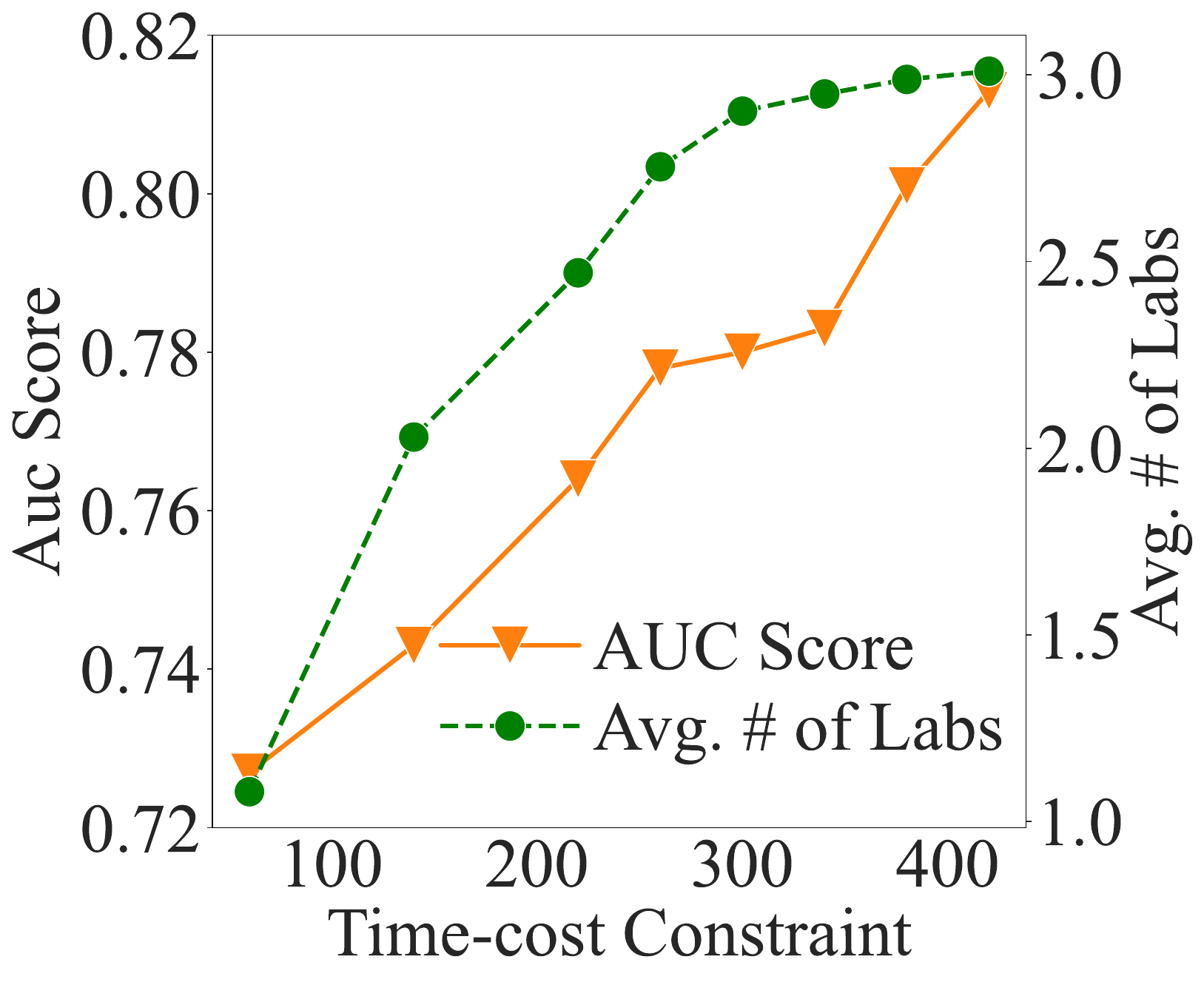}
        \caption{Critical Outcome AUC}
        \label{fig:curve_critical_auc}
    \end{subfigure}%
    \begin{subfigure}[t]{0.25\textwidth}
        \centering
        \includegraphics[width=\textwidth]{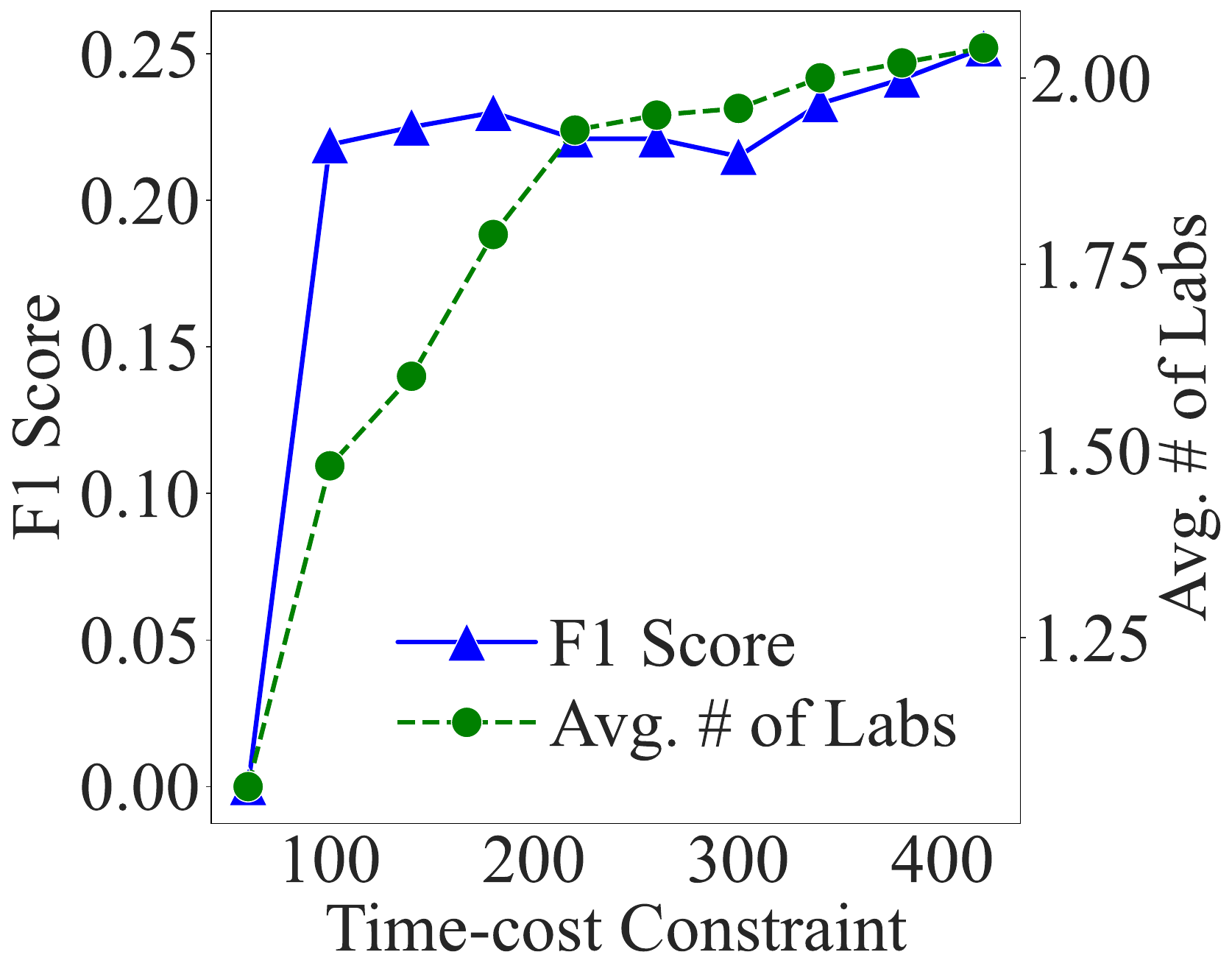}
        \caption{Lengthened ED Stay F1}
        \label{fig:curve_los_f1}
    \end{subfigure}%
    \begin{subfigure}[t]{0.25\textwidth}
        \centering
        \includegraphics[width=\textwidth]{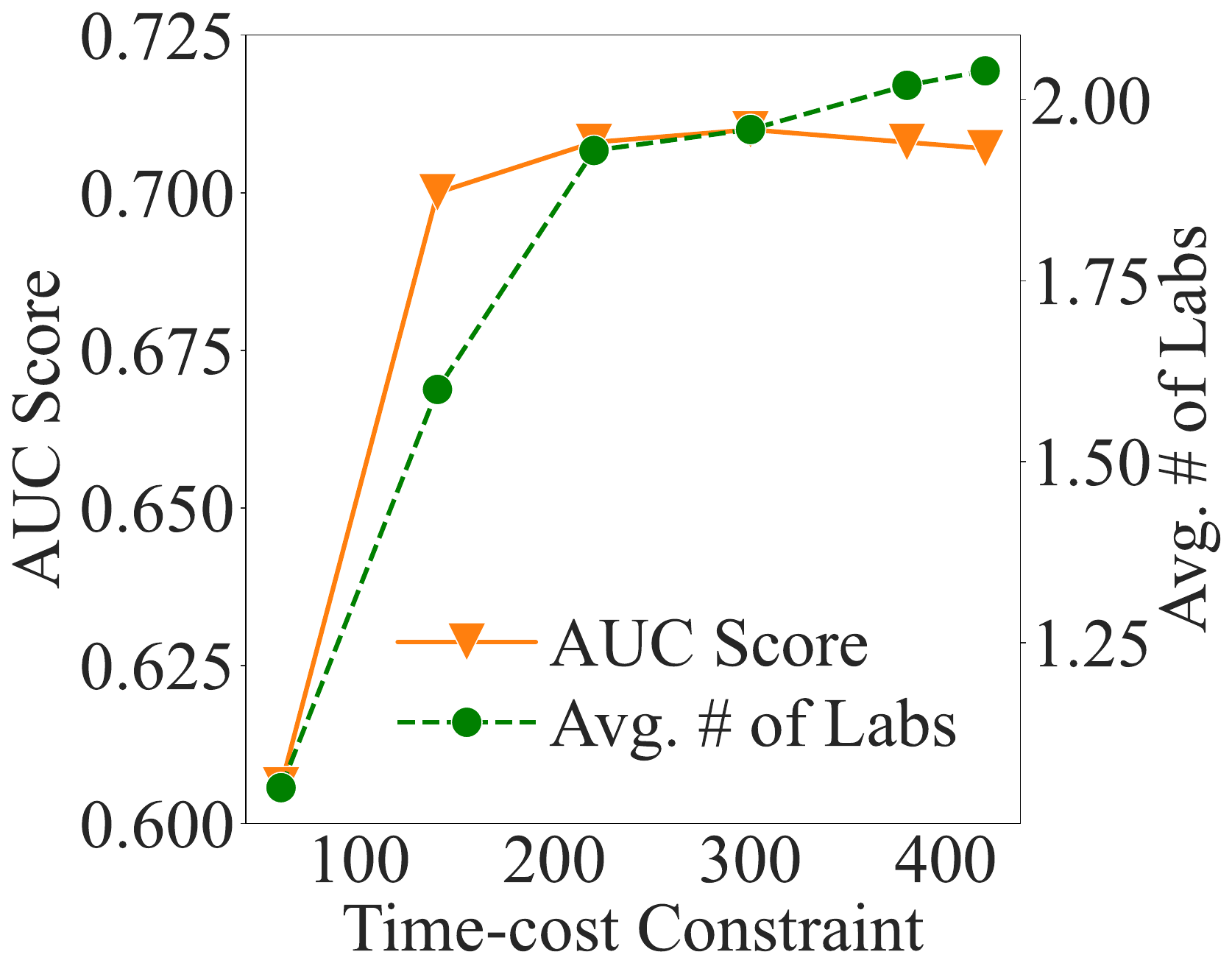}
        \caption{Lengthened ED Stay AUC}
        \label{fig:curve_los_auc}
    \end{subfigure}
    \vspace{-0.5cm}
    \caption{Prediction accuracy and average number of laboratory groups of ED-Copilot with different maximum allowed time to perform laboratory tests. Each point reflects \method's F1/AUC (y-axes) at different time upper-bounds.}
    \label{fig:cost_curve}
\end{figure*}

\section{Experimental Set-up}\label{sec:experimental_set_up}

\textbf{Dataset Split.} We randomly split the dataset using $80\%$ for training, $10\%$ for validation, and $10\%$ for testing, while ensuring each split has the same class distribution.
The validation set is used to tune hyper-parameters. 
During inference, the initial state of a patient is set to their triage information.

\textbf{Evaluation Metrics.} We use four evaluation metrics: F1-score, area under the receiver operating characteristic (AUC), sensitivity (true positive rate), and specificity (true negative rate) which are 
all standard in healthcare tasks prediction tasks~\cite{Harutyunyan_2019,xie2022benchmarking}.

\textbf{Baselines.} For prediction tasks only, ED-Copilot is compared to three tree-based methods: random forests \cite{breiman2001random},  XGBoost \cite{chen2016xgboost} and LightGBM \cite{ke2017lightgbm}, all of which are known to achieve strong performance on tabular data. We also compare ED-Copilot to a $3$-layer dense deep neural network (DNN).

Since each patient is not observed under all laboratory groups, there are many missing values in \benchmark.
Since baselines cannot naturally handle missing values, we consider a variety of imputation methods: (1) mean imputation, (2) median imputation, and (3) zero-imputation (i.e., replacing missing values by $0$), (4) using a dummy indicator to encode missing values. 
Results for the best imputation method are presented in the main manuscript. The rest are in Appendix~\ref{sec:imputation_results}. 

We compare \method\ to another cost-effective baseline, SM-DDPO~\cite{yu2023deep}, which selects laboratory groups to minimize time-cost while maximizing F1-score. 
SM-DDPO is a non-personalized method, i.e., it selects the same laboratory groups for all patients.

\textbf{Time-cost.} As discussed, each group of tests is assigned a time-cost by observing time-stamps in the MIMIC-IV database. 
For all methods, we measure the amount of time taken by averaging the time required for an algorithm to make a prediction across patients. 

\textbf{Implementation Details.} BioGPT \cite{luo2022biogpt} is used as our backbone language model, which is a generative pre-trained transformer (347 million parameters) for biomedical text generation and representation. 
In the RL phase of training ED-Copilot, for a given patient, our action space is restricted to laboratory groups that the patient actually performed. 
Experiments when \method\ is allowed to select unobserved laboratory groups can be found in \cref{sec:free_generation}.

%


%

%

\begin{table*}[h]
  \centering
\caption{Ablation study on components of ED-Copilot, including linearization techniques, feature importance, and PLM backbones.}
\resizebox{\textwidth}{!}{
  \begin{tabular}{llccccrccccr}
    \toprule
    \multirow{2}{*}{\textbf{Group}} & \multirow{2}{*}{\textbf{Variations}} & \multicolumn{5}{c}{\textbf{Critical Outcome}} & \multicolumn{5}{c}{\textbf{Lengthened ED Stay}} \\
    \cmidrule(lr){3-7} \cmidrule(lr){8-12}
    & & F1 & AUC& Sensitivity&Specificity & Avg. Time-cost & F1 & AUC& Sensitivity&Specificity & Avg. Time-cost \\
    \midrule
     &ED-Copilot (345M) & \textbf{0.413} &\textbf{ 0.820} & \textbf{0.750} & \textbf{0.779 }& \textbf{125 Min} & \textbf{0.252} & \textbf{0.707}&\textbf{0.725} &\textbf{0.606} & \textbf{154 Min} \\
    \midrule
    \textbf{Linearization} & Raw Lab Test Name & 0.397 & 0.777 & 0.768&0.677 & 134 Min& 0.241 & 0.695 & 0.611 & 0.701 & 144 Min \\
    \midrule
    \multirow{3}{*}{\textbf{Features}} & w/o. Triage & 0.277 & 0.704 & 0.679 & 0.649 &---  & 0.145 & 0.593 & 0.532 & 0.606 & --- \\
    & w/o. CBC & 0.385 & 0.803 & 0.692 & 0.777 & --- & 0.224 & 0.686  & 0.696 & 0.596 & --- \\
    & w/o. CHEM & 0.420 & 0.827 & 0.788 & 0.746 & --- & 0.234 & 0.702 &0.656 &0.606 & --- \\
    \midrule
    \multirow{4}{*}{\textbf{Backbone}} 
    & BioGPT (345M) w/o. RL  & 0.381  & 0.810 & 0.725       & 0.765       & 265          Min & 0.236  & 0.718 & 0.710       & 0.620       & 265 Min \\
    & Llama (7B LORA) w/o. RL & 0.397  & 0.798 & 0.692       & 0.767       & 265       Min    & 0.232  & 0.701 & 0.705       & 0.610       & 265 Min \\
    \cdashline {2-12}
    & Pythia (70M) w. RL & 0.290 & 0.698 & 0.574 & 0.702 & 166 Min & 0.178 & 0.596 & 0.555 & 0.619 & 126 Min \\
    & GPT2-Medium (345M) w. RL & 0.358 & 0.757 & 0.621 & 0.747 &133 Min & 0.166 & 0.539 & 0.498 & 0.584 & 96 Min \\
    \bottomrule
  \end{tabular}
}
  \label{ablation}
\end{table*}



\section{Evaluation Results}
\label{sec:results}
In this section, we present our experimental results.
Section \ref{sec:prediction_time_cost} compares the predictive and time-cost performance of \method{} with baselines. 
Section \ref{sec:ablation} describes our ablation studies. 
Sections \ref{sec:personalization} and \ref{sec:subgroup} establish \method's ability to personalize recommendations based on patient severity  and achieve consistent performance across subgroups such as age and sex. 
Lastly, in Section \ref{sec:free_generation}, we investigate \method's performance when it is not restricted to select administered groups. 

\subsection{Prediction Accuracy and Time-cost}
\label{sec:prediction_time_cost}
\textbf{Overall Performance.} 
The prediction performance and time-costs of all methods are presented in \cref{major}. 
For critical outcome, \method{} outperforms the next best baseline by 1.9\%, 0.7\%, 2.5\%, and 1\% for F1-score, AUC, sensitivity, and specificity, respectively.  
For lengthened ED stay, \method{} outperforms the next best model by 3.5\%, 0.2\%, 1.9\%, and 0.1\% for the four metrics. 
\method{} halves average time-costs from roughly $4$ hours to $2$ hours.

In comparison to the other cost-effective baseline, SM-DDPO, \method{} also has significantly better accuracy and lower time-costs. 
Both methods utilize the same RL algorithm to minimize time-cost, but \method{} benefits from the strong capability of PLM backbone (Sec~\ref{sec:ablation}) and the personalized diagnostic assistance it enabled (Sec~\ref{sec:personalization}).

%
%

%

\textbf{Prediction Accuracy at Different Time-costs.} In \cref{fig:cost_curve}, we investigate how \method's\ prediction performance changes as we vary the maximum time allowed to perform laboratory tests. 
Performance increases with maximum allowed time and number of laboratory groups selected. 
\method{} requires relatively few tests to achieve peak performance for lengthened ED stay, while prediction accuracy for critical outcome steadily increases as more tests are allowed.
These curves illustrate different requirements for different outcomes and shed light as to how clinicians should allocate resources across tasks.

\subsection{Ablation Studies}
\label{sec:ablation}

\textbf{Linearization Technique.} 
Following \citet{hegselmann2023tabllm} and \citet{NEURIPS2023_9f94298b}, we replace the true laboratory test name with a standard feature name while keeping the result for that laboratory group unchanged.
That is, we linearize laboratory results as follows ( \textit{feature1 : value1 $|$  feature2 : value2 ... }). 
\cref{ablation} shows using standard feature names leads to better F1 and AUC score as compared to using the raw laboratory test names.
One possible reason is that using raw laboratory test names biases \method{} to pick names over informativeness for prediction.

\textbf{Feature Importance.} 
We examine the change in \method{} performance  
when training with and without a given feature set from top three popular feature sets: triage information, complete blood count (CBC), and chemical group (CHEM). 
Table \ref{ablation} shows removing triage features leads to a significant drop in performance. 
Our ED collaborators confirms this is medically plausible since initial laboratory tests are based on triage information.
Our results also show all laboratory groups are not equally important, for example, removing CBC causes a larger drop in performance than removing the chemical laboratory group. 

\begin{figure}[t]
    \centering
    \begin{subfigure}[t]{0.25\textwidth}
        \centering
        \includegraphics[width=\textwidth]{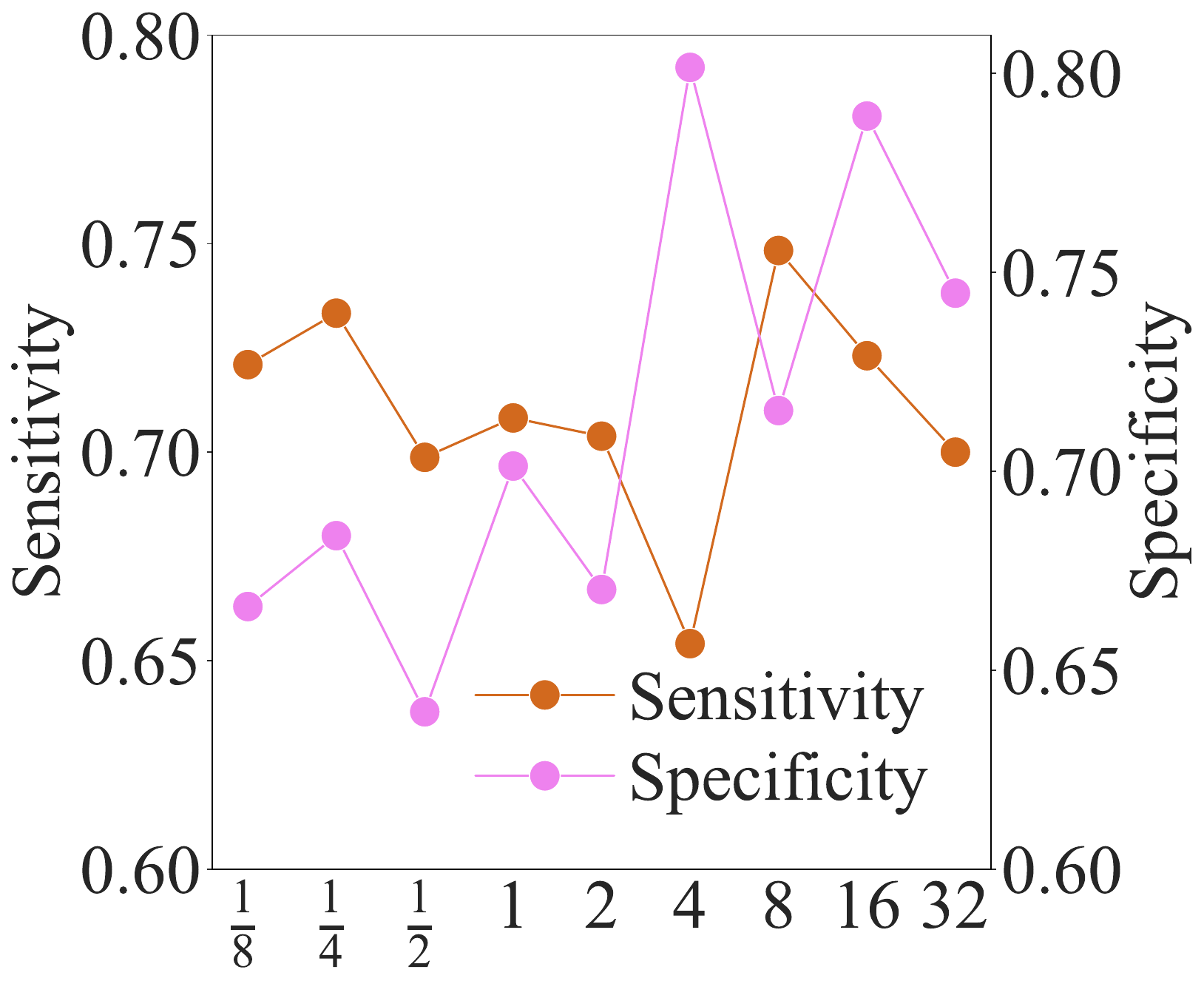}
        \caption{Sensitivity-Specificity $\alpha$}
        \label{fig:reward_critical_sens_spec_alpha}
    \end{subfigure}%
    \begin{subfigure}[t]{0.25\textwidth}
        \centering
        \includegraphics[width=\textwidth]{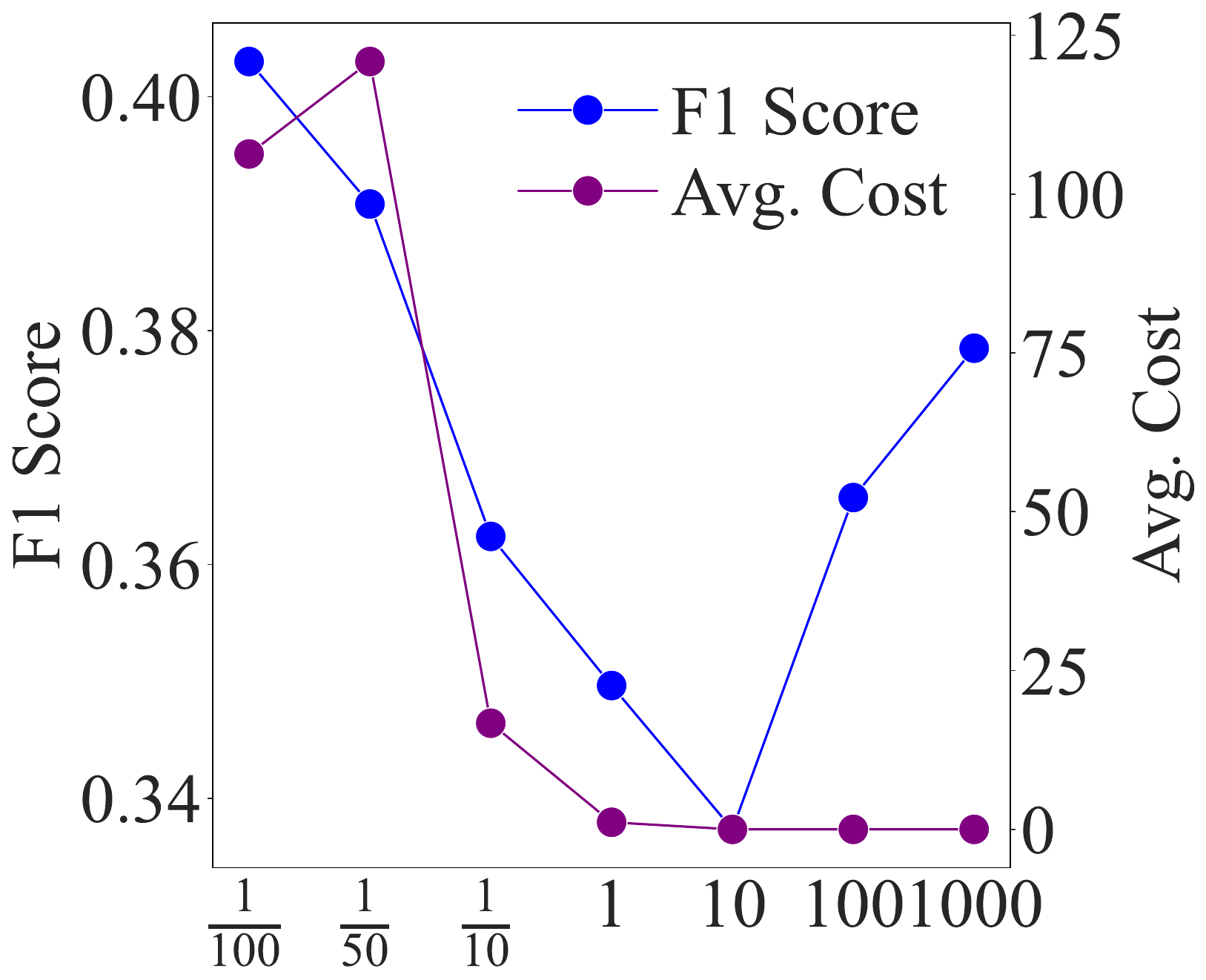}
        \caption{F1-Cost $\beta$}
        \label{fig:reward_critical_f1_cost_beta}
    \end{subfigure}
      \vspace{-0.3cm}
    \caption{Impact of Hyper-parameters on Sensitivity-Specificity ($\alpha$) and F1-Cost ($\beta$) trade-off when predicting critical outcome.
    }
    \label{fig:balance}
\end{figure}

\textbf{PLM Backbone.} 
We compare BioGPT to GPT2 which is the same size, and a smaller model Pythia \cite{biderman2023pythia} that has 70 million parameters. 
BioGPT is based on GPT2 \cite{radford2019language}, but is further pre-trained on a bio-medical corpus.
\cref{ablation} shows both model scale and bio-medical pre-training lead to stronger performance.  
We also compare Llama (7B) \cite{touvron2023llama} without RL for this task to BioGPT (345M) without RL, and ED-Copilot.
Llama is fine-tuned using Low-Rank Adaptation (LORA) \cite{hu2022lora} for efficiency.
Llama 7B (LORA) without RL performs comparably to BioGPT without RL.
This indicates that scaling up the backbone parameters is promising avenue to boost absolute performance.
However, Llama 7B (LORA) underperforms BioGPT with RL (i.e., ED-Copilot), particularly when it comes to time-cost.
This is expected as the PLM backbone and the RL serve different roles in the system.
PLM is used to predict outcomes while RL minimizes time-cost.
Fine-tuning does not address time-costs. These two aspects are both necessary.

\textbf{Hyper-parameters.} 
Hyper-parameters $(\alpha,\beta)$ in \cref{eq:optimal_policy} control the trade-off between prediction accuracy and time-cost in training.
Increasing $\alpha$ trades off sensitivity over specificity, while increasing $\beta$ trades off F1-score over time-cost.
Figure~\ref{fig:reward_critical_sens_spec_alpha} shows increasing $\alpha$ leads to trade off sensitivity over specificity.
Figure \ref{fig:reward_critical_f1_cost_beta} shows increasing $\beta$ causes \method{} to trade off F1-score over time-cost.
The hyper-parameters $(\alpha,\beta)$ provide a control for ED clinicians to change \method's behavior to fit their preferences.
\begin{figure}[t]
    \centering
        \centering
        \includegraphics[width=0.485\textwidth]{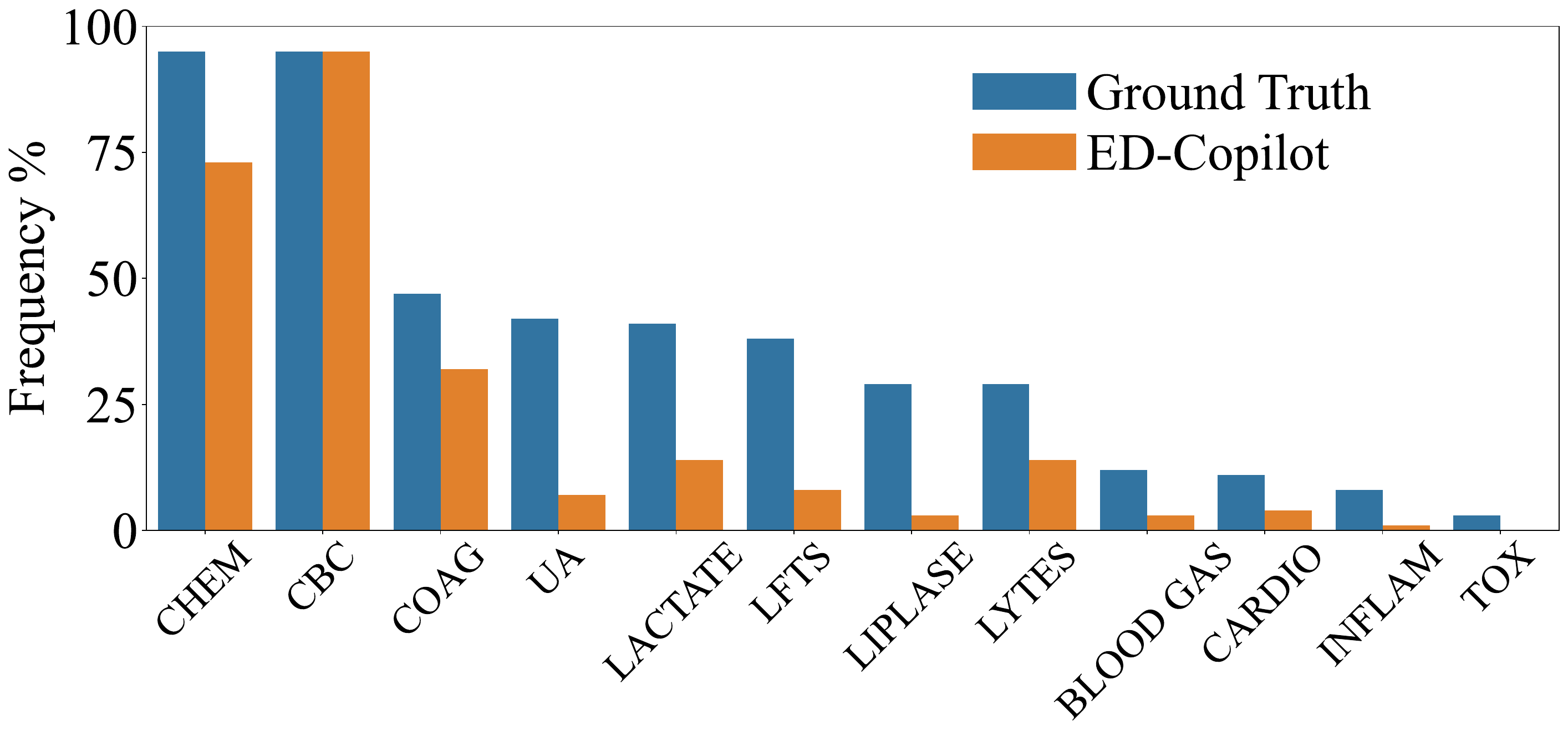}
        \vspace{-0.5cm}
   \caption{Fraction of patients performing laboratory groups and predicted by ED-Copilot. On average each patient performed 4.7 groups and cost-effective ED-Copilot suggested 2.4 groups.}
    \label{fig:personalization}
\end{figure}




\begin{table*}[h]
\caption{Critical Outcome prediction accuracy at different cohorts, by the category of laboratory groups they performed (Top: Performed at least one of the top two most frequent laboratory groups; Middle: Performed at least one of the middle six laboratory groups; Rare: Performed one or more rare tests in the last four laboratory groups). The total number of positive (critical)/negative cases and positive rate is shown in parentheses. 
}
\label{tab:personalization}
\centering
\resizebox{\textwidth}{!}{
\begin{tabular}{lccccccccc}
\toprule
& \multicolumn{3}{c}{\textbf{W. Top 2 Lab Groups} (302/2823,9.6\% )} & \multicolumn{3}{c}{\textbf{W. Middle 6 Lab Groups} (299/2603, 10.3\%)} & \multicolumn{3}{c}{\textbf{W. Last 4 Lab Groups} (141/817, 14.7\%)} \\
\cmidrule(lr){2-4} \cmidrule(lr){5-7} \cmidrule(lr){8-10}
\textbf{Model} & F1 & Sensitivity & Specificity & F1 & Sensitivity & Specificity & F1 & Sensitivity & Specificity \\
\midrule
Random Forest (Top 2 groups) & 0.330 & 0.735 & 0.752 & 0.335 & 0.746 & 0.750 & 0.405 & 0.730 & 0.739 \\
XGBoost (Top 2 groups) & 0.361 & 0.788 & 0.680 & 0.374 & 0.732 & 0.728 & 0.433 & 0.738 & 0.718 \\
LightGBM (Top 2 groups) & 0.401 & 0.788 & 0.705 & 0.409 & 0.806 & 0.690 & 0.462 &  0.738 & 0.742 \\
\midrule
SM-DDPO  &  0.364 & 0.760 & 0.715 & 0.373  &0.762& 0.724&0.435& 0.732&0.734\\
ED-Copilot & 0.414 & 0.701 & 0.788 & 0.431 & 0.731 & 0.774 & 0.461 & 0.767 & 0.720 \\
\bottomrule
\end{tabular}
}
\end{table*}



\begin{table}[h]
\large
  \centering
\caption{Model performance using only triage information under supervised fine-tuning training.}

\resizebox{0.48\textwidth}{!}{
\begin{tabular}{lcccccccc}
\toprule
\multirow{2}{*}{\textbf{Model}} & \multicolumn{4}{c}{\textbf{Critical Outcome}} & \multicolumn{4}{c}{\textbf{Lengthened ED Stay}} \\
\cmidrule(lr){2-5} \cmidrule(lr){6-9}
                       & F1    & AUC   & Sensitivity & Specificity & F1    & AUC   & Sensitivity & Specificity \\ 
\midrule
Random Forest          & 0.346 & 0.767 & 0.689       & 0.731       & 0.181 & 0.617 & 0.597       & 0.571       \\ 
XGBoost                & 0.324 & 0.736 & 0.673       & 0.668       & 0.165 & 0.603 & 0.580       & 0.560       \\ 
LightGBM               & 0.341 & 0.743 & 0.624       & 0.738       & 0.171 & 0.615 & 0.554       & 0.628       \\ 
3-layer DNN            & 0.275 & 0.712 & 0.611       & 0.702       & 0.155 & 0.553 & 0.485       & 0.597       \\ 
\midrule
ED-Copilot (SFT)       & 0.392 & 0.792 & 0.708       & 0.762       & 0.199 & 0.675 & 0.679       & 0.557       \\ 
\bottomrule
\end{tabular}%
}
  \label{tab:only_triage_train}
\end{table}

\subsection{Analysis on Personalized Diagnostic Assistance}
\label{sec:personalization}
Personalized treatment strategies are often required since patients have diverse needs and present differently. 
\method{} learns a personalized representation for each patient and naturally supports personalized diagnostic assistance. 
Previous cost-effective methods based on traditional ML algorithms such as tree-based methods optimize cost-effectiveness at the full population level, e.g., via feature selection. 
Unlike \method{}, this does not lead to personalized recommendations that are based on triage and previous test results. 
The following experiments highlight the benefit of personalization via \method{}.

We plot both the fraction of patients receiving each group of tests as well as the fraction of patients predicted by \method{} to undergo each of these groups in \cref{fig:personalization}.
This plot shows significant variation in tests received by
ED patients. 
After the two most common groups (CHEM and CBC), more than half of the patients performed some other tests.
\method{} reflects this diversity by recommending a variety of tests based on patients' condition. 
On the other hand, non-personalized cost-effective methods often withdraw to picking the most frequent groups. 
%

To further understand the benefit of personalized diagnostic assistance, we partition patients in the test set into three cohorts based on the rarity of laboratory groups they were administered. 
A precise definition of these cohorts can be found in \cref{tab:personalization}. 
The middle and rare cohorts have a larger proportion of positive cases, indicating higher severity. 
We train tree-based baselines using triage information and results from the two most frequent groups (CHEM and CBC), and compare it to \method. 
Results in \cref{tab:subgroup} show \method{} and baselines perform comparably on patients in the low severity cohorts (i.e., patients that present typically).
For the middle cohort, \method{} greatly improves F1-score. 
In the rare cohort which has the highest proportion of critical outcomes (15\% versus 10\% in middle cohort), \method{} achieves higher significantly higher sensitivity than other methods. 
This indicates \method's ability to provide personalized assistance which results in more 
equitable care, particularly for high-risk patients.


\begin{table*}[h]
  \centering
    \small
\caption{Sub-group Analysis for Critical Outcome and Lengthened ED Stay. }
\resizebox{\textwidth}{!}{
\begin{tabular}{lcccccccccc}
\toprule
\multirow{2}{*}{\textbf{Model}} & \multicolumn{5}{c}{\textbf{Critical Outcome}}& \multicolumn{5}{c}{\textbf{Lengthened ED Stay}} \\
\cmidrule(lr){2-6} \cmidrule(lr){7-11}
& F1 & AUC & Sensitivity & Specificity & Avg. Time-cost & F1 & AUC& Sensitivity&Specificity & Avg. Time-cost \\
\midrule
 \multicolumn{11}{l} {\textbf{Group}: Male, \textbf{Positive/Negative (Ratio)}: 155/1329 (10.44\%) on critical outcome, 104/1380 (7.01\%) on lengthened ED stay}\\
\midrule
Random Forest & 0.387 & 0.793 & 0.785 & 0.692 & 265 &0.207 & 0.723 & 0.644       & 0.704       & 265    \\
XGBoost & 0.379 & 0.804 & 0.726 & 0.732 & 265 &  0.258 & 0.722 & 0.692       & 0.659       & 265 \\
LightGBM & 0.391 & 0.807 & 0.762 & 0.742 & 265 & 0.223 & 0.724 & 0.769       & 0.597       & 265  \\
ED-Copilot & 0.377 & 0.780 & 0.671 & 0.775 & 121 & 0.242 & 0.670 & 0.574       & 0.702       & 151          \\
\midrule
  \multicolumn{11}{l} {\textbf{Group}: Female, \textbf{Positive/Negative (Ratio)}: 143/1608 (8.17\%) on critical outcome, 127/1624 (7.25\%) on lengthened ED stay}\\
\midrule
 Random Forest & 0.385 & 0.819 & 0.730 & 0.794 & 265 & 0.193 & 0.677 & 0.646       & 0.616       & 265 \\
XGBoost & 0.368 & 0.826 & 0.766 & 0.740 & 265 & 0.171 & 0.645 & 0.661       & 0.576       & 265    \\
LightGBM & 0.394 & 0.813 & 0.766 & 0.730 & 265 & 0.198 & 0.683 & 0.669       & 0.636       & 265    \\
ED-Copilot & 0.413 & 0.823 & 0.727 & 0.819 & 129 & 0.215 & 0.694 & 0.713       & 0.621       & 160 \\
\midrule
  \multicolumn{11}{l} {\textbf{Group}: Age 18-30, \textbf{Positive/Negative (Ratio)}: 22/341 (6.06\%) on critical outcome, 23/340 (6.34\%) on lengthened ED stay} \\
\midrule
Random Forest & 0.277 & 0.871 & 0.889 & 0.838 & 265 & 0.197 & 0.742 & 0.739       & 0.738       & 265 \\
XGBoost & 0.413 & 0.908 & 0.833 & 0.884 & 265 & 0.306 & 0.764 & 0.652       & 0.824       & 265\\
LightGBM & 0.390 & 0.895 & 0.889 & 0.913 & 265 & 0.259 & 0.767 & 0.783       & 0.629       & 265  \\
ED-Copilot & 0.208 & 0.849 & 0.864 & 0.708 & 106 & 0.304 & 0.823 & 0.956       & 0.705       & 159      \\
\midrule
  \multicolumn{11}{l} {\textbf{Group}: Age 31-60, \textbf{Positive/Negative (Ratio)}: 97/1169 (7.66\%) on critical outcome, 92/1174 (7.27\%) on lengthened ED stay}\\
\midrule
Random Forest & 0.367 & 0.824 & 0.739 & 0.794 & 265 & 0.196 & 0.675 & 0.620       & 0.670       & 265\\
 XGBoost & 0.367 & 0.835 & 0.750 & 0.775 & 265 & 0.228 & 0.685 & 0.674       & 0.658       & 265  \\
LightGBM & 0.400 & 0.835 & 0.802 & 0.756 & 265 & 0.213 & 0.692 & 0.739       & 0.611       & 265   \\
ED-Copilot & 0.409 & 0.800 & 0.711 & 0.823 & 124 & 0.209 & 0.697 & 0.612       & 0.664       & 153 \\
\midrule
  \multicolumn{11}{l} {\textbf{Group}: Age 61-90, \textbf{Positive/Negative (Ratio)}: 189/1261 (13.03\%) on critical outcome, 108/1342 (7.45\%) on lengthened ED stay} \\
\midrule
 Random Forest & 0.411 & 0.780 & 0.710 & 0.753 & 265 & 0.208 & 0.707 & 0.685       & 0.621       & 265 \\
 XGBoost & 0.383 & 0.785 & 0.659 & 0.789 & 265  & 0.188 & 0.655 & 0.704       & 0.552       & 265   \\
LightGBM & 0.405 & 0.779 & 0.688 & 0.757 & 265 & 0.192 & 0.698 & 0.667       & 0.622       & 265\\
ED-Copilot & 0.400 & 0.749 & 0.667 & 0.694 & 129 & 0.222 & 0.695 & 0.731       & 0.671       & 156 \\
\bottomrule
\end{tabular}%
}
\label{tab:subgroup}
\end{table*}

\subsection{Sub-group Analysis}
\label{sec:subgroup}

We compare the performance of ED-Copilot to tree-based baselines across sub-groups including gender and age to study if the algorithm is working fairly across all sub-groups of interest in Table \ref{tab:subgroup}. ED-Copilot displays consistent performance, and outperforms baselines across subgroups on both prediction tasks.

\subsection{Performance of Unrestricted Lab Group Suggestion}
\label{sec:free_generation}
Since \benchmark\ is an offline retrospective benchmark, we restrict \method{} during training to only select laboratory groups that patients have received. 
However, this restriction may lead to sub-optimal group suggestion.
We investigate \method{}'s performance without this restriction through the following experiment. 
\begin{figure}[t]
    \centering
     \begin{subfigure}[t]{0.22\textwidth}
        \centering
        \includegraphics[width=\textwidth]{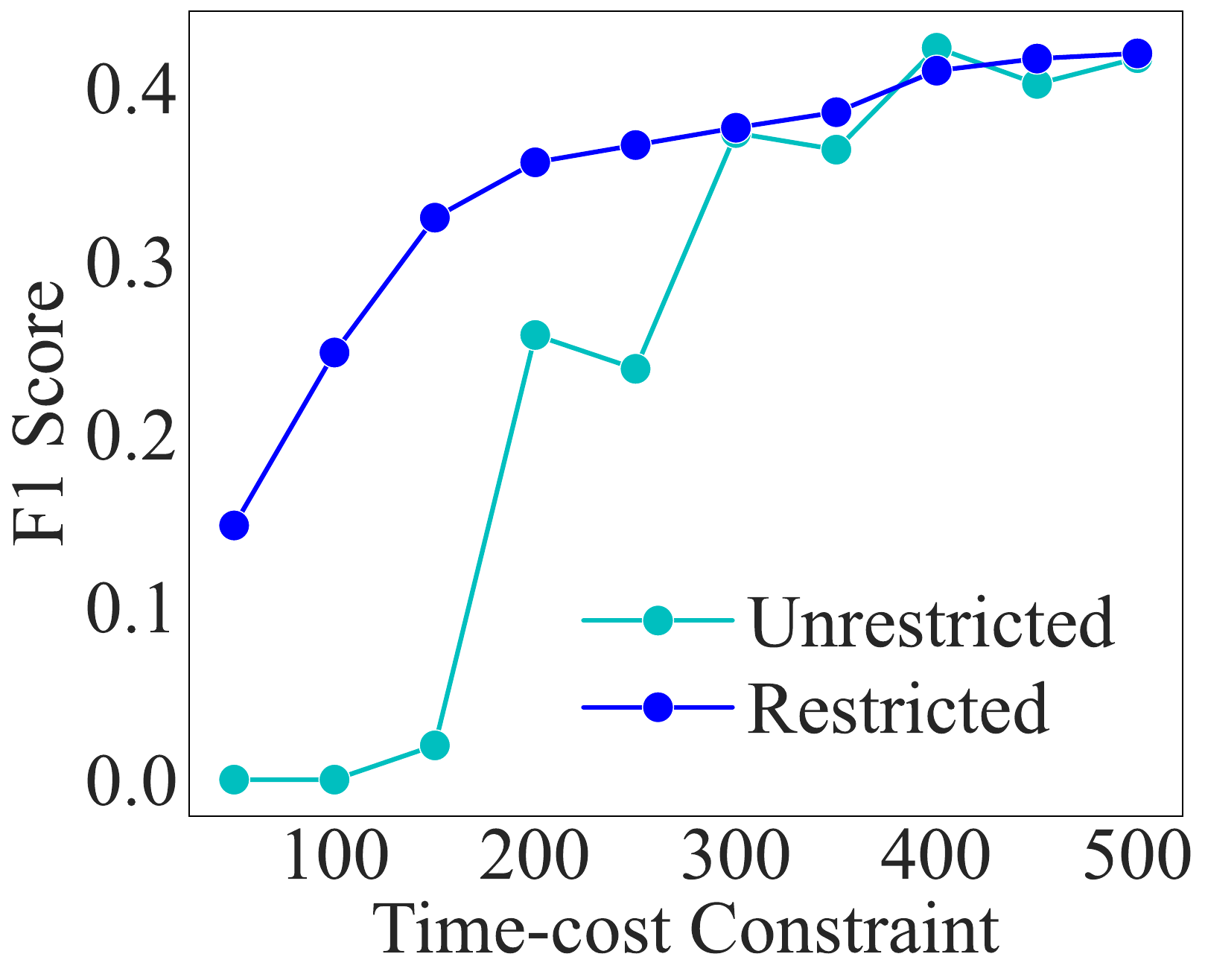}
        \caption{Accuracy}
    \end{subfigure}~
         \begin{subfigure}[t]{0.22\textwidth}
        \centering
        \includegraphics[width=\textwidth]{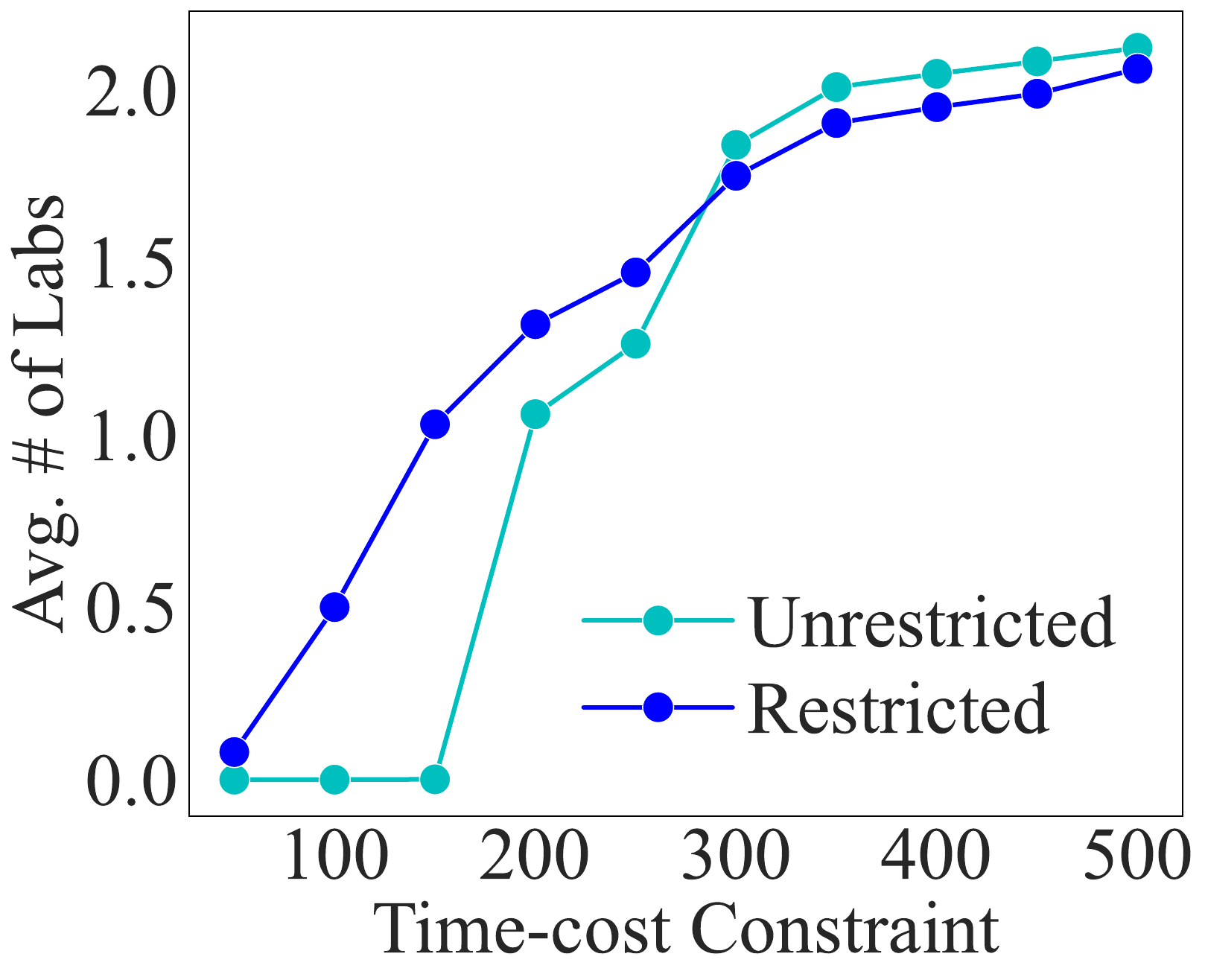}
        \caption{Avg. \# of Labs}
    \end{subfigure}
    \vspace{-0.3cm}
   \caption{Comparison of prediction performance on critical outcome of \method{}  when restricted or not restricted to tests patients performed in its suggestion at different time-cost constraints. 
   }
    \label{fig:freeform}
\end{figure}

If  \method{} (unrestricted) selects a laboratory group not received by a patient, we impute those results by $0$, but add to the time-cost. 
%
%
\method{} (unrestricted) achieves an accuracy of $77\%$ when predicting the next laboratory group, i.e., the group actually received by a given patient, indicating that \method\ (unrestricted) is reasonably good at replicating clinical decisions but there are still differences between model suggestions and historical data.

Next, we compare \method\ (unrestricted)  to \method{} (restricted) in terms of F1-score for critical outcome when varying the maximum allowed time. \cref{fig:freeform} shows that when maximum allowed time is small, \method{} (unrestricted) performs worse than \method{}. 
This is because if \method{} (unrestricted) picks a laboratory test that was not administered to a patient, this selection does not result in new information being added. 
However, as the number of allowed laboratory groups increases, \method{} (unrestricted) matches the performance of \method. 
Restriction to observed laboratory groups is necessary for offline evaluation.
In real-world settings, \method{} should serve as a ``Co-Pilot'' and suggest laboratory groups from the entire set without restriction, and assist ED clinicians in optimizing their workflows.

We also evaluate ED-Copilot on new patients only using triage data, and without any laboratory results which are the information available on patient arrival.
We use ED-Copilot only with supervised fine-tuning (SFT), and without RL.
Results are displayed in Table \ref{tab:only_triage_train}.
Without laboratory results, all models suffer, but ED-Copilot (SFT) outperforms baselines, showing its utility at an early treatment stage.

%

\section{Conclusion}
\label{sec:conclusions}

In this paper, we aim at reducing emergency department wait time through time-cost-effective diagnostic assistance. 
In collaboration with ED clinicians, we curate \benchmark, a benchmark that contains comprehensive laboratory information annotated with medically relevant groupings alongside key clinical outcomes that serve as useful approximations to ED wait-time. 
Using \benchmark, we propose \method, an AI system that provides time-cost-effective diagnostic assistance for ED clinicians, recommending informative laboratory groups and flagging patients at high risks of critical outcome and lengthened ED stay. 
\method\ outperforms baselines by significantly improved prediction accuracy and reduced laboratory time-costs.
Its language model backbone allows for personalized diagnostic assistance to better address the needs of severe patients. 
We believe this work takes a step towards AI-driven assistance in the ED and hope that \benchmark\ spurs interest in applying advancements in AI to tackle this critical healthcare challenge.

\section*{Acknowledgments}
We gratefully acknowledge partial support from NSF grants DMS-2209975, 2015341, NSF grant 2023505 on Collaborative Research: Foundations of Data Science Institute (FODSI), the NSF and the Simons Foundation for the Collaboration on the Theoretical Foundations of Deep Learning through awards DMS-2031883 and 814639, and NIH grant R01GM152718.
We also acknowledge support by the Eunice Kennedy Shriver National Institute of Child Health and Human Development of the National Institutes of Health under Award Number K23HD110716. 
We also thank Yikuan Li from Northwestern University to share their pre-processed sepsis dataset.

\section*{Impact Statement}
\label{sec:impact}
The development of \method\ marks an advancement in medical technology, offering an AI-driven diagnostic tool to improve patient care in ED. 
By expediting diagnoses, \method\ has the potential to increase efficiency of ED operations and enhance patient care. 

From an ethical perspective, the deployment of \method\ carries a significant responsibility on the privacy and security of patient data. As the system will handle sensitive health information, strict measures, for example, internal-accessible only systems and compliant data protection procedures, must be in place to protect against breaches and misuse, ensuring patient confidentiality.

In the broader societal context, the implementation of \method\ aims to address the issue of ED congestion. It is vital to ensure that the benefits of such technology are accessible to all segments of society, regardless of socioeconomic status. Equity in healthcare technology means that all patients, irrespective of their background, should have the opportunity to benefit from advancements like \method. Moreover, the development of \method\ should support healthcare professionals, not replace them, and should be viewed as a tool to assist medical staff, allowing them to focus on the more nuanced aspects of patient care.

\bibliography{reference}
\bibliographystyle{icml2024}

\newpage
\appendix
\onecolumn

\section{Triage and Laboratory Test Group }
\label{sec:triage_lab_features}
In Table \ref{tab:lab},  we include triage assessments and 12 laboratory groups with 68 tests and their estimated time-costs.

\begin{table}[h]
    \caption{Triage and Lab Test Group }
    \centering
\resizebox{1\textwidth}{!}{
    \begin{tabular}{l|l|l}
    \hline
    \multirow{7}{*}{Triage} 
    & Age &  Diastolic blood pressure \\
    & Gender   & Self-reported pain \\
    & Heart rate & Emergency severity index acuity\\
    & Respiratory rate     & Chief complaint \\
    & Systolic blood pressure & Oxygen saturation  \\
    & Temperature & Past commodities \\
    & Past ICU/ED/hospital visiting frequency &  \\
    \hline
    \multirow{12}{*}{Complete Blood Count (CBC) (30 min)} & Hematocrit   & Neutrophils \\
    & White Blood Cells  & Red Cell Distribution Width (Standard Deviation)  \\
    & Hemoglobin  & Absolute Lymphocyte Count \\
    & Red Blood Cells  & Absolute Basophil Count  \\
    & Mean Corpuscular Volume  & Absolute Eosinophil Count \\
    & Mean Corpuscular Hemoglobin  & Absolute Monocyte Count \\
    & Mean Corpuscular Hemoglobin Concentration  & Absolute Neutrophil Count  \\
    & Red Blood Cell Distribution Width  & Bands  \\
    & Platelet Count & Atypical Lymphocytes \\
    & Basophils  & Nucleated Red Cells\\
    & Eosinophils  & Monocytes \\
    & Lymphocytes \\
      \hline
    \multirow{5}{*}{Chemistry (CHEM) (60 min)} & Urea Nitrogen & Glucose (Chemistry) \\
    & Creatinine & Potassium \\
    & Sodium   & Anion Gap \\
    & Chloride  & Calcium, Total \\
    & Bicarbonate \\
    \hline
    \multirow{2}{*}{Coagulation (COAG)(48 min)} & Prothrombin Time  & Partial thromboplastin time  \\
    &  International Normalised Ratio \\
    \hline
    \multirow{5}{*}{Urinalysis (UA) (40 min)} & PH (Urine) 
 & Protein  \\
    & Specific Gravity   & Hyaline Casts \\
    & Red Blood Count (Urine)   & Ketone \\
    & White Blood Count (Urine) & Urobilinogen  \\
    & Epithelial Cells    & Glucose (Urine) \\
    \hline
    \multirow{1}{*}{Lactate (4 min)} & Lactate \\
    \hline
    \multirow{3}{*}{Liver Function (LFTs)(104 min)} & Alkaline Phosphatase  & Bilirubin, Total \\
    & Asparate Aminotransferase (AST)  & Albumin \\
    & Alanine Aminotransferase (ALT) \\
    \hline
    \multirow{1}{*}{Lipase (100 min)} & Lipase \\
    \hline
    \multirow{1}{*}{Electrolyte (LYTES)(89 min)} & Magnesium   & Phosphate  \\
    \hline
    \multirow{1}{*}{Cardiovascular (CARDIO) (122 min)} & NT-proBNP   & Troponin T \\
    \hline
    \multirow{4}{*}{Blood Gas (12 min)} & Potassium, Whole Blood  & PO2  \\
    & PH (Blood Gas) & PCO2  \\
    & Calculated Total CO2   & Glucose (Blood Gas) \\
    & Base Excess   & Sodium, Whole Blood \\
    \hline
    \multirow{1}{*}{Toxicology (TOX) (70 min)} & Ethanol \\
    \hline
    \multirow{1}{*}{Inflammation (INFLAM) (178 min)} & Creatine Kinase (CK) & C-Reactive Protein \\
    \hline
    \end{tabular}}
    \label{tab:lab}
\end{table}

\section{Training Details}
\label{sec:training}
We use a $3$-layer neural network for all MLPs in \method. 
Due to large class imbalance, we use class weights when training the diagnostic predictor $p_{\psi}$.
%
All experiments, training and hyper-parameter tuning are conducted on one NVIDIA RTX A6000 GPU. 
During the RL phase of training \method, we restrict the action space 
to be on laboratory groups that were administered to patients. 
That is, we do not select laboratory groups that patients have not received. 
%
%
To ensure sufficient policy sampling on experience replay buffer and obtain a large batch size, we freeze the weights of our PLM.
To train the policy $\pi_{\eta}$, we use the PPO algorithm \cite{schulman2017proximal}. The loss function employed is as the following for hyper-parameters $c_1$ and $c_2$: 
\begin{align*}
\mathcal{L}_{\text{rl}} (\eta; \eta^{old}) &= \mathcal{L}_{\text{clip}}(\eta; \eta^{old}) - c_1 \mathcal{L}_{\text{value}}(\eta) + c_2 \text{Entropy}[\pi_{\eta}],\\
\mathcal{L}_{\text{clip}}(\eta; \eta^{old}) &= {\hat{\mathbb{E}}}_i \left[ \min\left(\frac{\pi_{\eta}(a_i | {\textbf{\textit{h}}}_{<i})}{\pi_{\eta^{old}}(a_i | {\textbf{\textit{h}}}_{<i})} \hat{A}_{i}, \text{clip}\left(\frac{\pi_{\eta}(a_i | {\textbf{\textit{h}}}_{<i})}{\pi_{\eta^{old}}(a_{i} | {\textbf{\textit{h}}}_{<i})}, 1 - \epsilon, 1 + \epsilon \right) \hat{A}_{i}\right) \right],\\
\mathcal{L}_{\text{value}}(\eta) &= \hat{\mathbb{E}}_{i} \left[ (V_{\eta}({\textbf{\textit{h}}}_{<i}) - V_{i})^2 \right].
\end{align*}
Here ${\textbf{\textit{h}}}_{<i}$ is the hidden representation from PLM of $\texttt{[EOS]}_{<i}$ in this patient's linearized sequence in state $s_i$.
$\mathcal{L}_{\text{clip}}$ is clipped surrogate loss and $\hat{A}_{i}$ is estimated advantages which are regularized by value function $\mathcal{L}_{\text{value}}$. 
$\text{Entropy}[\pi_{\eta}]$ denotes an entropy term over the states, and $\hat{\mathbb{E}}_i$ is the empirical average over the collected dataset. 
We used a masked actor-critic network with package stable-baseline3 \cite{stable}.
See algorithm \ref{alg:ppo} for a high-level description of this method. 
\begin{algorithm*}[t]
\caption{Proximal Policy Optimization (PPO)}
   \label{alg:ppo}
\begin{algorithmic}
\FOR{iteration \( = 0,1,\ldots \)}
    \FOR{actor \( = 1,2,\ldots,N_{\text{actor}} \)}
        \STATE Run policy \( \pi_{\eta_{\text{old}}} \) in environment for \( T \) timesteps and save all observations in experience buffer
        \STATE Compute estimated advantage  \( \hat{A}_1,\ldots,\hat{A}_{i} \)
    \ENDFOR
    \STATE Optimize surrogate \( \mathcal{L}_{\text{rl}} \) w.r.t \( \eta \), with \( D \) epochs and minibatch size \( M \leq N_{\text{actor}}T \)
    \STATE \( \eta_{\text{old}} \leftarrow \eta \)
\ENDFOR
\end{algorithmic}
\end{algorithm*}

\begin{table*}[t]
  \centering
  \caption{Hyperparameters Configurations for ED-Copilot}
  \label{tab:hyper}
\resizebox{1\textwidth}{!}{
\begin{tabular}{lcc}
\toprule
\textbf{Hyperparameter} & \textbf{Supervised Fine-tuning} & \textbf{Reinforcement Learning} \\
\midrule
Learning Rate & \multicolumn{2}{c}{1e-5} \\
DNN Hidden size (3-layer) & \multicolumn{2}{c}{1024} \\
Optimizer & \multicolumn{2}{c}{AdamW} \\
Adam $\epsilon$ & \multicolumn{2}{c}{1e-8} \\
Adam Betas ($\beta_1$, $\beta_2$) & \multicolumn{2}{c}{(0.9, 0.999)} \\
Weight decay & \multicolumn{2}{c}{0.01} \\
Batch Size & 32 & 128 \\
Epochs & 15 & 10 \\
Max sequence length & 656 & 656 \\
Class Weight & 10 & \multicolumn{1}{c}{-} \\
Warmup percentage & 0.1 & \multicolumn{1}{c}{-} \\
Buffer Steps & \multicolumn{1}{c}{-} & 2048 \\
Timesteps PPO trained per epoch & \multicolumn{1}{c}{-} & 20000 \\
The penalty ratio between false positive and false negative $\alpha$ & \multicolumn{1}{c}{-} & 15 \\
The penalty ratio between false positive and laboratory cost  $\beta$  & \multicolumn{1}{c}{-} & $\frac{1}{100}$\\
\bottomrule
\end{tabular}}
\end{table*}
We list the hyper-parameters in Table \ref{tab:hyper}, including the supervised fine-tuning and reinforcement learning stage. In the RL stage, we use grid-search to tune $\alpha$ and $\beta$ to balance the trade-off between accuracy and cost. The search scope for $\alpha \in \{\frac{1}{8},\frac{1}{4},\frac{1}{2},1,2,4,8,15,16,32,64,256\}$ and $\beta \in \{\frac{1}{100},\frac{1}{50},\frac{1}{20},\frac{1}{10}, 1,10,100,1000 \}$. The parameters in PPO that are not specified are assigned the default values found in the Python package  \cite{schulman2017proximal}. 

\newpage
\section{More Baseline Results}
\label{sec:imputation_results}

In Table \ref{tab:imputation}, we present tree models' results with different imputation methods: (1) Mean imputation. (i.e. replacing missing values with the mean value of the non-missing data for the particular feature). (2) Median imputation. (i.e. replacing missing values with the median value of the non-missing data for the particular feature) (3) Zero imputation. (i.e. replacing missing values by 0). (4) A dummy indicator to encode missing values, which is used in XGBoost and LightGBM.

\begin{table*}[h]
  \centering
\caption{Results with Different Imputation Methods}
\resizebox{1\textwidth}{!}{
  \begin{tabular}{llcccccccc}
    \toprule
    \multirow{2}{*}{Model} & \multirow{2}{*}{\textbf{Method}} & \multicolumn{4}{c}{\textbf{Critical Outcome}} & \multicolumn{4}{c}{\textbf{Lengthened ED Stay}} \\
    \cmidrule(lr){3-6} \cmidrule(lr){7-10}
    & & F1 & AUC& Sensitivity&Specificity & F1 & AUC& Sensitivity&Specificity \\
    \midrule
    \multirow{3}{*}{Random Forest} & Mean & 0.377 & 0.807 & 0.754 & 0.748  & 0.198 & 0.701 & 0.736 & 0.578  \\
    & Median & 0.355 & 0.807 & 0.793 & 0.709  & 0.206 & 0.698 & 0.693 & 0.616  \\
    & Zero & 0.367 & 0.808 & 0.738 & 0.748  & 0.194 & 0.702 & 0.628 & 0.674  \\
    \midrule
    \multirow{4}{*}{XGBoost} & Mean & 0.328 & 0.768 & 0.715 & 0.697   & 0.212 & 0.678 & 0.593 & 0.683  \\
    & Median & 0.358 & 0.783 & 0.731 & 0.708 & 0.197 & 0.674  & 0.658 & 0.601  \\
    & Zero & 0.374 & 0.818 & 0.735 & 0.754  & 0.212 & 0.680 &0.619 &0.662  \\
    & Dummy & 0.379 & 0.807 & 0.731 & 0.744  & 0.190 & 0.656 &0.680 &0.557 \\
    \midrule
    \multirow{4}{*}{LightGBM} & Mean & 0.390 & 0.812 & 0.764 & 0.732  & 0.217 & 0.705 & 0.707 & 0.606  \\
    & Median & 0.394 & 0.813 & 0.725 & 0.769 & 0.190 & 0.681  & 0.688 & 0.599  \\
    & Zero & 0.393 & 0.814 & 0.735 & 0.774  & 0.209 & 0.702 &0.697 &0.623  \\
    & Dummy & 0.379 & 0.807 & 0.731 & 0.744  & 0.192 & 0.671 &0.654 &0.629  \\
    \bottomrule
  \end{tabular}
}
  \label{tab:imputation}
\end{table*}


\end{document}